\documentclass[runningheads]{llncs}
\usepackage{caption}
\usepackage{enumitem}
\usepackage{amsmath}
\usepackage{graphicx}
\usepackage{subfigure}
\begin{document}
\title{Car Monitoring System in Apartment Garages by Small Autonomous Car using Deep Learning}
%\thanks{National University of %Engineering \& Center Information %Technology and Communications.}

%
\titlerunning{Monitoring System by Mini Robot}
\author{Leonardo Le\'on-Vera\inst{1} 
\and Felipe Moreno-Vera\inst{1}
\and Renato Castro-Cruz\inst{1}
\and Jose Nav\'io-Torres\inst{1}
\and Marco Capcha-Mansilla\inst{2} \\
\email{\{lleonv, felipe.moreno.v, racastroc, jnavio\}@uni.pe and cm.marco919@gmail.com}
}
\authorrunning{L. Le\'on et al.}

\institute{Universidad Nacional de Ingenier\'ia \inst{1}\\
\url{http://www.uni.edu.pe}\\
Universidad Tecnol´ogica del Per\'u \inst{2}\\
\url{https://www.utp.edu.pe/}}

\maketitle  % typeset the header of the contribution

\begin{abstract}
Currently, there is an increase in the number of Peruvian families living in apartments instead of houses for the lots of advantage; However, in some cases there are troubles such as robberies of goods that are usually left at the parking lots or the entrance of strangers that use the tenants parking lots (this last trouble sometimes is related to kidnappings or robberies in building apartments). Due to these problems, the use of a self-driving mini-car is proposed to implement a monitoring system of license plates in an underground garage inside a building using a deep learning model with the aim of recording the vehicles and identifying their owners if they were tenants or not. In addition, the small robot has its own location system using beacons that allow us to identify the position of the parking lot corresponding to each tenant of the building while the mini-car is on its way. Finally, one of the objectives of this work is to build a low cost mini-robot that would replace expensive cameras or work together in order to keep safe the goods of tenants.

\end{abstract}

\section{Introduction}

\begin{figure}[htbp]
\centering
  \includegraphics[width=9cm]{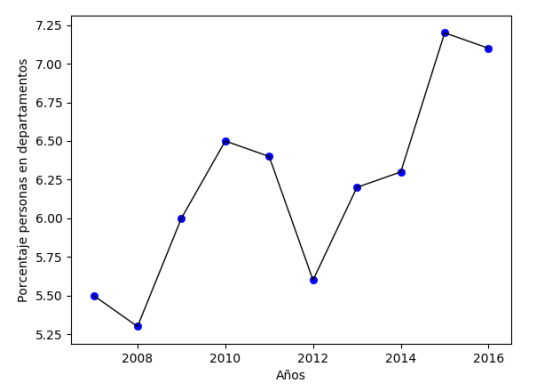}
  \caption{Growth of people living in apartments. Source: Private households of Peruvians 2007-2016. INEI.}
  \label{fig:correlation}
\end{figure}

Peruvian families prefer to live in apartments (see Fig.\ref{fig:correlation}), since the common advantages are more comfortable spaces and less effort in cleaning unnecessary rooms; however, the size of the apartments is reduced every year. Therefore, families decide to keep their goods at parking lots next to their vehicles or inside of them, as long as they are in an underground garage, but some of these stuff are stolen. Also, people who do not live in the buildings use the parking lots that are only for building users, sometimes thieves parked in the building garage before a stealing inside the building.

A smart self-driving mini-car is proposed to supervise the vehicles in the underground garage by identifying their owners using computer vision in their respective parking lots during a schedule that does not bother the tenants and cars. Using autonomous driving techniques and object detection, a computational vision system is proposed in order to achieve the autonomous navigation of the mini-car and the identification mainly of the cars owners with an ALPR (Automatic License Plate Recognition) based on a deep learning model that was developed by the open source organization OpenALPR.

Also, the mini-car has its own location system with Beacons that determine its relative position in real time and it can verify if a vehicle would be at its correct parking lot or not. Vehicles of the future will be strongly dependent on the development of the software algorithms controlling the autonomous navigation and how they can process the massive amount of data generated, so this mini-car wants to be a prototype of that type of vehicle that is able to be autonomous and at the same time translate the license plates that recognize into an important information for security purposes.

The navigation of the mini car in \cite{autonomous_car} is based on images recognition using deep learning, developed for more complicated tracks than an underground garage and it uses different sensors for better accuracy in indoor scene.

\section{Structure of the System}

\begin{figure}[htbp]
\centering
  \includegraphics[width=11cm]{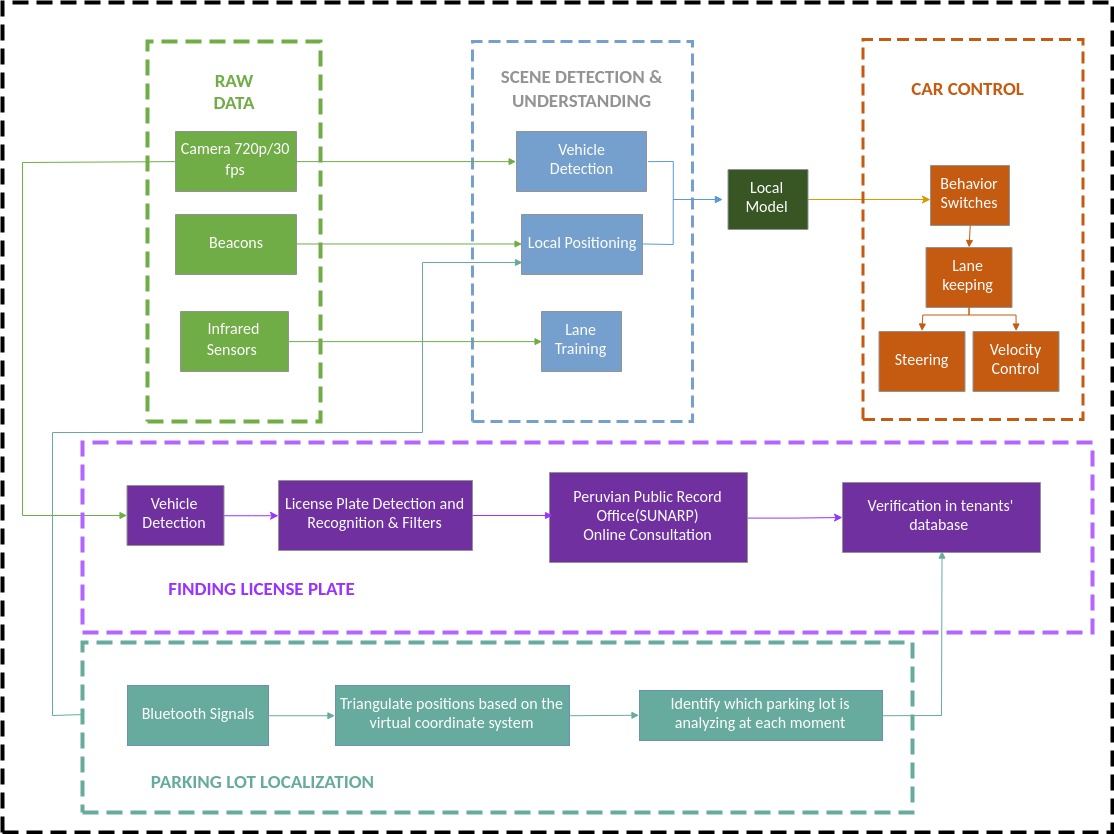}
  \caption{Structure of the system joining each task.}
  \label{fig:structure}
\end{figure}

Implementation of a mini-robot that runs through an underground garage and verify that the tenants cars are in their respective parking lots. It is proposed to solve the problem by detecting objects to identify the vehicle position and its license plate to obtain information from the owner of the car. The holistic system is made up of three main parts: The computer vision system that allows the self-driving and the license plate recognition; the location system that allows us to know the mini-car location and which parking lots are near at the same time; and the system that describes how the mechanic and electronic mini-car components are integrated (See Fig.\ref{fig:structure}).

\subsection{Data collected from tenants of the building}

\begin{figure}[htbp]
\centering
  \includegraphics[width=9cm]{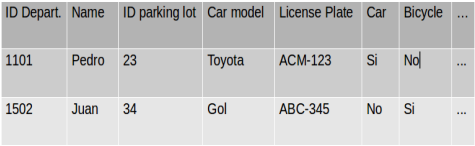}
  \caption{Data of tenants.}
  \label{fig:dataset}
\end{figure}

The first step to build a well-designed system that verifies if the vehicle parked on its respective parking lot is having the owner information. This data is collected from the tenants of the building such as the apartment number, name, ID of parking lot, type of vehicle, license plate, objects they stored in the parking lot, etc (See Fig.\ref{fig:dataset}). The owners name and their license plates will be compared with the data obtained by the mini-car license plate recognition and shown to the caretaker so that he can draw his conclusions if a bizarre event happens.

\subsection{Autonomous Navigation of the small car}

To face this task we use logitech C920 web cam which is cheaper than Lidars and Radars. These are very used for self-driving training since they can calculate the distance between objects with high precision. In contrast, it turns out that web cams emulate the way in which people can see the environment giving a better classification and interpretation of images textures in comparison to the previous ones \cite{radar_lidar}. We are imitating Behavioral Cloning method described in Udacity Self Driving Course because there is only one way where the mini car can go. In other words, It only has to turn left when it is going down and turn right when it is going up. It is one constraint about the garage.

\textbf{Data Collection}: Being a supervised learning, it was necessary to collect a video recording the route that the mini car must follow to obtain an optimal performance. Being a flat floor, we did not get additional problems; however, when the garage walls are very similar, it is needed to look for a correct camera position and cut out the image so that it only focus on taking images that correctly identify curves of straight roads. The input data are the images of three front cameras (images from one camera and 2 images modified) separated from each other by few centimeters to collect more data from the road and speed of the small car each moment; and angles between 0 and 180 as output values for training the model.\\

\textbf{Neural Network Architecture}: A deep sequence model of layers is used in the following manner, with its respective number of filters: Conv24-Conv36-DropoutConv48-Dropout-Conv64-Dropout-Conv64-Dropout-FC-FC-FC-FC with non - linear activation function ReLU, which bring good results in computational vision tasks in the Convolutional layers. Regularization methods such as Dropout with a probability of 0.5 are added, due to the overfitting that occurs with the NVIDIA model when iterations are increased considerably.\\

\textbf{Neural Network Training and Testing}: The training of the data was done with the board NVIDIA P4000 with a partition of 20\% validation set and 80\% training set. The MSE loss function and the ADAM optimizer method are used without the need to manually set the speed of the learning rate. Testing of the CNN is performed on the Udacity simulator, obtaining a clear outstanding performance when driving autonomously as can be seen in the video \cite{udacity_simulator} and later applied to Jetson TX1 in an underground garage.

\subsection{Mini-robot mechanism}

\begin{figure}[htbp]
\centering
  \includegraphics[width=8cm]{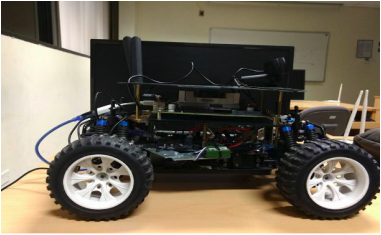}
  \caption{Mini-Robot for testing in the garage}
  \label{fig:minirobot}
\end{figure}

Mini-robot is built with base of a Monster Truck 1/10 which includes a motor and 2 servos for the movement of front and rear tires. An Arduino UNO is added that allows communication between the engine and the NVIDIA Jetson TX1 board (see in Fig. \ref{fig:minirobot}). The Arduino sends instructions to the engine through its GPIOs where indicates speed of the wheels and angle of rotation of the front wheels. Communication of the board TX1 to the Arduino is serial through a USB which allows to send integer values encoded in characters to indicate the rotation of the front wheels and the instructions to go forward, stop and rewind. Through Python program on the TX1 board, instructions are sent to the Arduino and this sends to the motor which angle have to turn.

\subsection{Vehicles Detection}

The first attempt to find objects in each parking slot was cars detection. A convolutional neuronal network is in charge of this task using Tiny YOLO model \cite{yolo_tiny} that is the model with the best performance between accuracy and inference time in detection objects in real time such as cars (our priority), bicycles and others objects trained using VOC dataset of 2012. The model also detects void spaces if probability to find any object is lower than the threshold. Once the car is detected, its position is saved in jetson’s memory with relative position brought by the beacons.\\

\textbf{Neural Network Architecture} Tiny YOLO model consists in a convolutional neuronal network with 9 convolutional layers of 16, 32, 64, 128, 256, 512, 1024, 512, 425 filters each one. This model is lightweight in comparison with ”YOLO9000: Better, Faster, Stronger” \cite{yolo_model} and accors.\\

\textbf{Neural Network Test} Test of the neural network was performed on an NVIDIA P4000 GPU server with images taken in the garage and on the NVIDIA Jetson TX1 board which we used in the mini-robot in the garage with a C920 camera obtaining 15 fps which works without any problem with the tiny YOLO model.

\subsection{License Plate Recognition}

After trying cars detection and getting acceptable results, OpenALPR library has been tried to detect the plate license of each car and later to compare with SUNARP dataset. The mini robot records videos on its way in the garage so that the license plate recognition could be done. After obtaining many candidates, we are going to select the license plate that have six digits (Peruvian’s license plate rule) and the major often weighted.

\begin{figure}[htbp]
\centering
  \includegraphics[width=9cm]{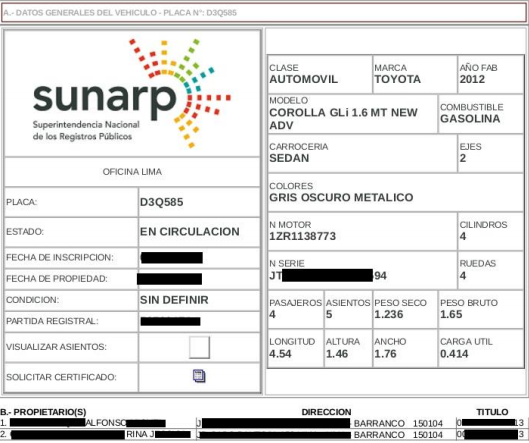}
  \caption{Query of the owner of license plate in SUNARP online application}
  \label{fig:sunarp}
\end{figure}

After recognizing the characters (letter and digits) of the license plates, the system must access the Peruvian public records office (SUNARP) online query \cite{sunarp} to collect information related to the owner of the vehicle (see Fig. \ref{fig:sunarp}). Once important information such as name and license plate is obtained, it will be compared with the database that the mini robot stores from collecting data of the building tenants.

\subsection{Tenant Parking Mapping by Beacons}

In this section, we present the way that robot knows where is it in real time. This action can be regulated with Beacons that is useful for indoor location in places where the Internet is not enough to make a connection \cite{indoor_location}, even allowing tracking in real-time applications \cite{real_time_indoor}.\\

\textbf{Detection of Beacons} In the garage, there is a distance about 8 meters between walls and 2.45 or 7.35 meters between vehicles, as shown in the Fig. 6(a) and Fig. 6(b).

\begin{figure}[htbp]
\centering
  \includegraphics[width=10cm]{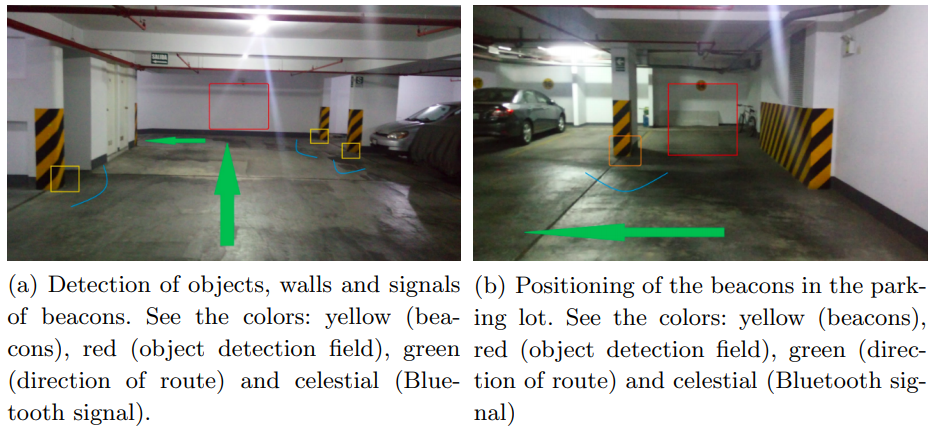}
  \caption{Environment and Beacon’s location map}
  \label{fig:distances}
\end{figure}

\begin{figure}[htbp]
\centering
  \includegraphics[width=7cm]{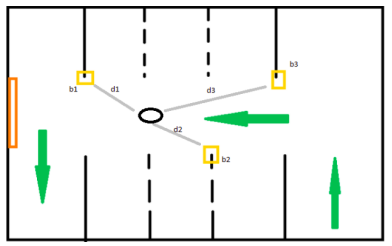}
  \caption{Locations of the beacons in the garage and variables to be taken for the
calculation and identification of the beacons.}
  \label{fig:beacons}
\end{figure}

As it is seen, Beacons have been distributed efficiently (see Fig.\ref{fig:beacons}) that allows to calculate where it is located in real time, as well as identifies who owns that area due to the match between image recognition and the current position, based on previous work \cite{beacons}.

\textbf{Determination of the position of the mini-robot} This section is about the equations that define the system to solve, by forming imaginary circles around each beacon, the respective radii are calculated. Then we proceed to calculate the relative position with respect to the global system of beacons based on distance equations forming a system of equations with the following form:
$$
E_i: (x - x_i)^2 + (y - y_i)^2 = d^2 _i
$$
Where (x, y) is the current position of the mini-robot and the indexes i corresponding to each beacon and $r_1
, ..., r_3$ to the respective vectors.

For this system formed proceeds to solve: Take ($x_i$, $y_i$) as coordinates of each beacon, we deduce $r_i = r_c + d_i$. Where $r_c$ is the current position of the mini-robot, for all equations the pair
of indices i, j are not equal (i $\neq$ j). So, taking the module we got:
$$
|r_i|^2 = |r_c|^2 + 2(r_c)(r_i) + |d_i|^2.
$$
Then, we calculate:
$$
|r_i|^2 - |r_j|^2 = 2 (r_c) (r_i - r_j) + |d_i|^2 - |d_j|^2
$$
We know  then we have:
$$
Y_i =  2 (r_c) (r_i - r_j) = |d_i|^2 - |d_j|^2 + |r_i|^2 - |r_j|^2
$$

Which is the same as: 
$$
Y_i = x_c (x_i - x_j) + y_c (y_i - y_j)
$$

Which forms a new linear system that is solved by numerical methods and
represented as:
$$
Y=AX
$$
Where:\\
Y: is the column vector of substractions between beacons.\\
X= $(x_c, y_c)^t$ is the column vector of the mini-robot positions\\
A=$\begin{bmatrix}
    x_1 − x_2 & y_1 − y_2 \\
    x_2 − x_3 & y_2 − y_3 \\
    x_3 − x_1 & y_3 − y_1 \\
\end{bmatrix}
$

\section{Results}

\subsection{Determination of the position of the mini robot}

To solving the Linear Equation System, we compared different methods, like iterative and directly. In directly methods we have the problem that the order of the algorithm is almost $O(n^3)$ like Gauss Elimination and derivate should be applied to special matrix (like symmetric, semi positive, positive, dominant
diagonal, etc).

Using Methods like Gauss Elimination and iterative methods like Jacobi and Gauss-Seidel (first verify if is a diagonal dominant matrix to apply) iterative methods), we parallel using OpenMP and CUDA. Using Montercalo methods to get an execution average time and approximate a pseudo real solution time.

\begin{table}[htbp]
    \begin{center}
    \begin{tabular}{|c|c|c|c|c|}
    \hline
    \multicolumn{4}{|c|}{\textbf{Result time to calculate current location}} \\
    \hline
    Method & Serial & Parallel (CPU=8) & GPU (Jetson)\\
    \hline
    Gauss-Seidel &  0.0729176 & 0.0152195 & 0.0000512\\
    \hline
    Jacobi & 0.366089 & 0.0151156 & 0.0000586\\
    \hline
    Gauss & 68.8792 & 17.0431 & 2.321465\\
    \hline
    \end{tabular}
    \end{center}
    \caption{Execution average time for 40000 times.}
    \end{table}

\subsection{Self-driving}

\begin{figure}[htbp]
\centering
  \includegraphics[width=8cm]{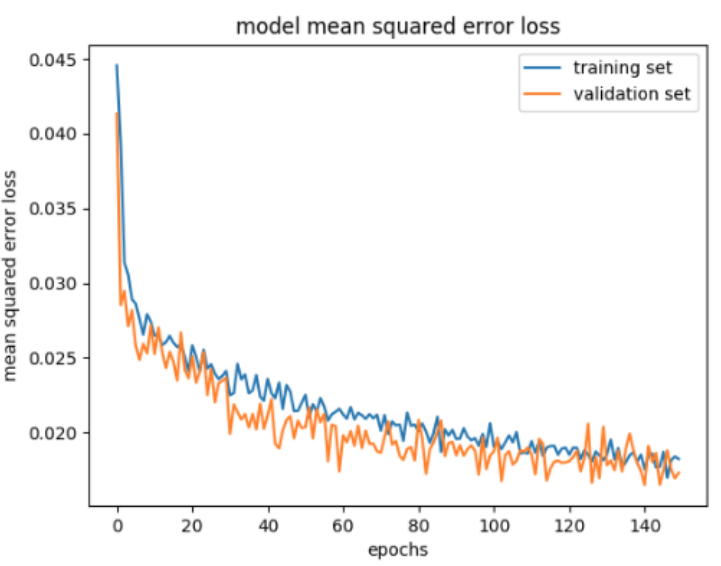}
  \caption{Graphics of demand and training}
  \label{fig:training}
\end{figure}

A Convolutional Neural Network based on the NVIDIA model \cite{nvidia_model} for regression is used to predict the angle of rotation of the wheels of the mini robot only obtaining an image of the path (See Fig. \ref{fig:training}). After training about 150 epochs we saw that the curve of loss function won’t be lower. We got 0.014 of Error in validation and training set using Udacity simulator. Next weeks, we are going to get results of our data obtained.

\subsection{Results of Cars Detection}

\begin{figure}[htbp]
\centering
  \includegraphics[width=8cm]{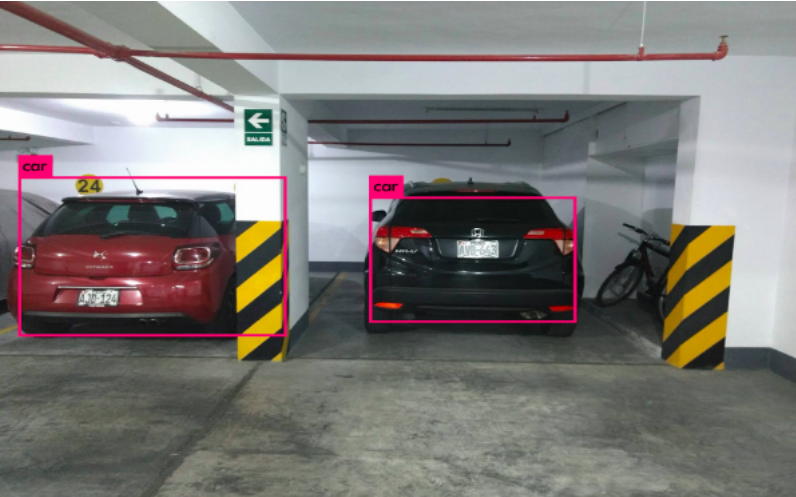}
  \caption{Detection of cars by mini robot}
  \label{fig:detection}
\end{figure}

Detection of objects is very well identifying cars of general classes (See Fig. \ref{fig:detection}), however when it is required to identify brand and model of the car there are no results since there is no dataset of images of front or back of the cars with annotations of their marks from which the neural network can learn. The nearest set of images found is Stanford dataset \cite{stanford_dataset}, but the majority of images are not taken from front or back view. The model of car recognition was trained with this dataset, but the images used for test were unlabeled. So, this dataset was discarded for the model.

\subsection{Results of Automatic License Plate Recognition}

OpenALPR library has a good performance in plate recognition, characters segmentation, and characters recognition \cite{plate_recognition}. OpenALPR has been trained on its own dataset, however there exists other datasets for this task such as SSIG dataset \cite{ssig_dataset}, a commercial dataset or UFPR-ALPR dataset \cite{yolo_plate_recognition} recently made with more fully annotated images and more vehicles in real-world scenarios for academic purposes.

\begin{figure}[htbp]
\centering
  \includegraphics[width=9cm]{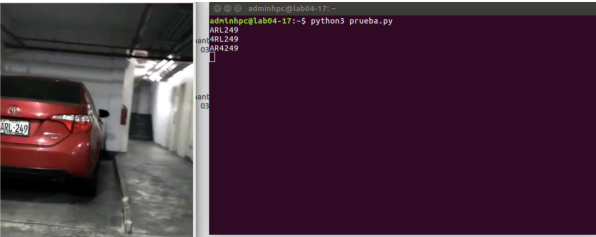}
  \caption{ License plate recognition of a video recorded by the mini-car}
  \label{fig:plates_recognition}
\end{figure}

\begin{figure}[htbp]
\centering
  \includegraphics[width=8cm]{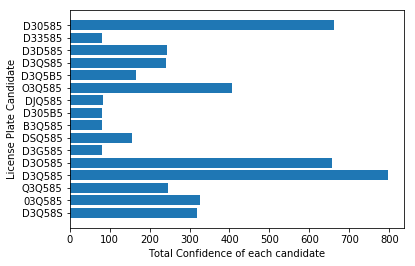}
  \caption{License Plates vs Total Confidence}
  \label{fig:plates_confidence}
\end{figure}

As it can be seen (See Fig. \ref{fig:plates_recognition}), the actual license plate was recognized with a high total confidence. It turns out that the input is a video file that was recorded by the mini-robot and its outputs in the console (See Fig. \ref{fig:plates_confidence}) are lots of candidates to be considered as the real license plate. After it, three candidates with the most high confidence are chosen to be verified in the tenants database and the SUNARP online consultation.

\section{Conclusions}

We see that during the implementation of the robot car, it has to count on many things that had not been foreseen to be able to handle the autonomous car.

In the work procedure, an optimal performance is obtained for the object detection and the automatic license plate recognition using Artificial Intelligence algorithms.
For the determination of positions of the mini robot based on the terrain delimited by columns and walls, it is necessary to specify the locations and the numerical system that solves the system of equations generated in such a way that a minimum error is obtained, which serves to determine and to identify the place where it is when we detect cars and objects of a certain car park, generating the relation object detected - location.

\end{document}